\documentclass[review]{elsarticle}

\usepackage{lineno,hyperref}
\usepackage{graphicx}
\usepackage{amsmath,amssymb} % define this before the line numbering.
\usepackage{bm}
\usepackage{color}
\usepackage[width=122mm,left=12mm,paperwidth=146mm,height=193mm,top=12mm,paperheight=217mm]{geometry}
\journal{Neurocomputing}

\begin{document}

\begin{frontmatter}

\title{Learning ordered pooling weights in image classification}


\author{J.I. Forcen}\corref{juan}
\address{Das-nano | Veridas,  31192, Tajonar, Spain \\
Universidad P\'ublica de Navarra \\ Campus Arrosad\'ia , 31006,  Pamplona, Spain}
\ead{jiforcen@das-nano.com}
\author{Miguel Pagola, Edurne Barrenechea}
\address{Universidad P\'ublica de Navarra \\ Campus Arrosad\'ia , 31006,  Pamplona, Spain}
\author{Humberto Bustince}
\address{Universidad P\'ublica de Navarra \\ Campus Arrosad\'ia , 31006,  Pamplona, Spain \\
King Abdullazih Universitiy, Jeddah, Saudy Arabia}

%\cortext[juan]{Corresponding author}

\begin{abstract}
Spatial pooling is an important step in computer vision systems like Convolutional Neural Networks or the Bag-of-Words method. The spatial pooling purpose is to combine neighbouring descriptors to obtain a single descriptor for a given region (local or global). The resultant combined vector must be as discriminant as possible, in other words, must contain relevant information, while removing irrelevant and confusing details.  Maximum and average are the most common aggregation functions used in the pooling step. To improve the aggregation of relevant information without degrading their discriminative power for image classification, we introduce a simple but effective scheme based on Ordered Weighted Average (OWA) aggregation operators. We present a method to learn the weights of the OWA aggregation operator in a Bag-of-Words framework and in Convolutional Neural Networks, and provide an extensive evaluation showing that OWA based pooling outperforms classical aggregation operators. Keras-TensorFlow implementation of ordered weighted pooling can be found at
https://github.com/jiforcen/ordered-weighted-pooling
\end{abstract}


\end{frontmatter}

%\linenumbers

%However, it does not encode global spatial information. Spatial pyramid model was introduced to incorporate loose spatial information.

\section{Introduction}

%A critical aspect in image classification is to find descriptive image features.

Image classification is one of the main problems in computer vision and pattern recognition, which plays an important role in scene understanding, object categorization, and many other vision tasks.

Pooling is an essential step in the state-of-the-art image classification methodologies as Convolutional Neural Networks (CNN) \cite{726791} or Spatial pyramid framework \cite{Lazebnik2006}. Pooling is an operation which aggregates local features into a new and more usable vector. The aggregation of local feature vectors produce more compact representations and improve robustness to noise, clutter and invariance to image translation and transformations.

%Moreover, there exist other classification frameworks like the VLAD \cite{Perronnin2012} or  the Super Vector  \cite{Zhou2010} that contain a pooling step. 
 
The result of a pooling operation depends on the aggregation operator $g$  used, because it defines how the local features are aggregated. The most common operators $g$ are the maximum, used in the well-known network architectures AlexNet \cite{AlexNet} or VGG \cite{VGG} and the arithmetic mean used in  Network in Network \cite{NiN} or GoogleNet \cite{GoogleNet}. Despite having a large influence on the performance of the network, most models use the maximum or the arithmetic mean \cite{Boureau2010}. The maximum and the mean have the advantage that, no parameters are learned in the pooling layers, however, the pooling operator $g$ used in each layer becomes an hyper-parameter of the network. Therefore, it is  another hyper-parameter of the network as the number of layers, number of filters per layer, the size of the filters, learning ratio, etc. It is well known that the selection, for a given problem, of the most appropriate network architecture and hyper-parameters has a high computational cost \cite{Bengio2012}.

Let's suppose a trained CNN with several layers, if the image $I$ which is being processed by the network contains, for example, a highly defined corner, it will produce a very high activation value in the feature channel where the corners are represented. If the maximum operator is used in the following pooling layer, said value (which is very high) will spread through the network architecture and become part of the final feature vector representation of the image. Therefore, if such corner is representative of the class of the image, then it will be a discriminative feature and will serve to correctly classify the image.

%On the other hand, we can think how this corner activation is propagated to the next step if average pooling is used, or we can think how affect both aggregations assuming that throughout the image $I$ are many corners similar to the case before but these are not clearly defined. If we think about this cases we will be aware that the activation propagation depends directly on the pooling operation chosen.

% The activation value of the pooled map will produce a final feature vector with also high activation of that feature if the pooling operator $g$ used is the arithmetic mean, but will be quite small compared with the previous case if max pooling is used.

Maximum pooling tends to give more importance to high activations, regardless of their frequency in the image, on the other hand the arithmetic mean filters the features that appear in isolation  and end up having a smaller value in the final representation. %However, average pooling  when combined with linear rectification non-linearities has the effect of down-weighting strong activations since many zero elements are included in the average.

The selection of the pooling operator $g$ is not a trivial decision because rely on each problem, and is influenced by factors like the images themselves, the features extracted from the images or the architecture proposed. 

%En este trabajo proponemos un método de pooling en el que se aprenda una función de {\it pooling}. El objetivo es aprender unos coeficientes que afecten únicamente a los valores de las activaciones (en el caso de la convolución los pesos están asociados a su posición espacial en el filtro). Para ello utilizaremos medias ordenadas ponderadas, en las que el vector de valores se ordena de mayor a menor y después cada valor es multiplicado por el coeficiente asociado a casa posición del vector ordenado. Los valores de los pesos se aprenderán en la fase de entrenamiento. De esta forma que nos proponemos obtener un operador de {\it pooling} que propague los valores más altos de las activaciones, pero también tenga en cuenta valores de activaciones medios que aparezcan frecuentemente. 

In this paper we propose a  new method called OWA-pooling, in which the aggregation is made by a weighted average of the ordered elements. In Ordered Weighted Aggregation functions, the weights are associated not with particular inputs, but with their magnitudes (in the case of a convolution the weights are associated to their spatial position in the filter). The weights will be learned in the training phase such a way the  pooling function will  propagate the highest values  activations, but also takes into account activation values that appear frequently.

In summary, the main contributions of this paper are the following ones:
\begin{enumerate}
\item We  propose to use Ordered Weighted Operators as the pooling operation in image classification methods.
\item We provide the methodology to learn the weights of the OWA-pooling operator such a way we obtain a parametrized transition between maximum and average pooling. 
\item We analyse the performance of the OWA-pooling in two different image classification frameworks: Bag-of-Words features with a global spatial pooling \cite{Lazebnik2006} and Convolutional Neural Networks. 
\item We show experimentally on public benchmarks that the proposed pooling method can provide performance gains with respect average pooling and maximum pooling.
\end{enumerate}

We noticed that Deliege et al. \cite{deliege2020ordinal} proposed a pooling methodology for CNNs based on sorting the pooling values called Ordinal Pooling \footnote{Both jobs were submitted in similar time, Deliège et. al. \cite{deliege2020ordinal} publication was made during the journal revision of this paper. We consider that both jobs are complementary, because of the implementation differences and as well as the wide experimentation presented in this job is different.}. The main difference between Deliege et al method and ours is that we keep the restrictions of the OWA operators, i.e. the sum of the weights must be one and postive. Also we developed the pooling strategy in traditional and modern computer vision frameworks and for local and global pooling in CNNs.

%Deliège et. al.  also worked in sort by magnitude the pooling values to aggregate , although in this work we use weights restrictions following OWA principles as was introduced in \cite{8015606}.

The remaining sections of this paper are organized as follows. Section \ref{sec:Relatedwork} contains a literature survey of related works. Section \ref{sec:BoW} presents a preliminary study of the proposed methodology in the Bag-of-Features methodology. In Section \ref{sec:CNN} the pooling methodology is described in CNNs. In both Sections \ref{sec:BoW} and \ref{sec:CNN} some experimental results are provided. We conclude in Section \ref{sec:conc}.

\section{Related work}\label{sec:Relatedwork}

Previous to the fast development of high-performance hardware and the great success of CNNs in image recognition,  Bag-of-Words (BoW) was the state of the art methodology in image classification \cite{Lazebnik2006}. BoW usually includes the following steps: feature detection, feature description known as coding, and pooling to obtain a final feature vector of the image. The final feature vectors are used to classify the image by means of a classifier, usually a Support Vector Machine (SVM). The process to obtain the final feature vector could be repeated on and on for regions of the image,  like the layers of a network \cite{Coates2011}, therefore in  \cite{Koniusz2013} and \cite{Boureau2010a} it is argued that the conclusions obtained for a pooling operator in a BoW model can be assumed in CNN models.  

In \cite{Boureau2010} and \cite{Boureau2010a} Boureau et. al. provide a  theoretical study and an empirical comparison between maximum and average pooling in the BoW model applied to image classification. They found that  maximum pooling is much better than average pooling, particularly, it is well suited for very sparse feature vectors. Furthermore, they prove that  for binary feature vectors, using the Bernoulli distribution under the i.i.d. assumption, the maximum expectation outperforms both average and maximum. The idea of using an expectation of the maximum comes from the fact that a negative factor of the pooling operation is to favour high values of local descriptors with low representativeness of the object of the image.  P. Koniusz et. al. \cite{Koniusz2013} try to solve this problem by simply pooling over the $n$ largest values.  Experiments carried out in \cite{Koniusz2013} use a fixed cardinality  $n$ of three, seven or fifteen. In \cite{IBPRIA2017} we studied the cardinality of the pooling operator that performs the average on the $\mathcal{N}$ major values (using BoW and Spatial Pyramid). Also in \cite{Pagola2017} we verified that the ordered weighted averages with fixed weights also give good results in this image classification methodology.

Regarding CNNs, Scherer et. al. \cite{Scherer2010} showed that a max pooling operation is superior compared to sub-sampling operations, however they found that max pooling easily overfits the training set.  In stochastic pooling \cite{Zeiler2013},  the pooled map response is generated by sampling from a multinomial distribution formed from the activations of each pooling region improving the performance in some datasets. In \cite{HUANG2014225} a global spatial structure is proposed with multiple Gaussian distributions which pools the features according to the relations between features and Gaussian distributions.

Due to hyper-parameter selection is a major drawback in CNNs, there are some works that propose to learn the pooling operator. For example, in \cite{ZHANG2018} is proposed the operator AlphaMEX which is a non-linear smooth log-mean-exp function to perform a global pooling before the softmax layer of a CNN. In \cite{Cui2017} was proposed a general pooling framework that captures higher order interactions of features in the form of a Gaussian Radial Basis Function kernel. Another  two methods are proposed in which the pooling function is learned  in \cite{pmlr-v51-lee16a}. In the first strategy,  a mixing parameter of the maximum value and the arithmetic mean value is learned. In the second method, a function of  pooling is learned in the form of a tree that mixes the results of different pooling filters. Sun et.al. \cite{SUN201796}  propose a learned pooling operation as a linear combination of the neurons in the region for each feature channel.

Basically, all of these methods try to find a method with the of max pooling and average pooling, but avoiding their drawbacks.

\section{Learning ordered weighted pooling weights in the Bag-of-Words setting}\label{sec:BoW}

In this section we present a first analysis of  learning weights in ordered weighted pooling in the BoW model for image classification. First, we recall the  
BoW method, that can be summarised in the following steps:

\begin{enumerate}
\item First, local image descriptors, such as SIFT \cite{Lowe2004}, HOG \cite{N.Dalal2005} or Gabor are extracted from images at interest points or in a dense grid. 
\item An unsupervised learning algorithm is used to discover a set of prototype descriptors that is called a dictionary. %This operation is usually done with k-means clustering, Gaussian mixtures or Sparse coding. Each of the descriptors found in the dictionary is called a visual word.
\item In the feature coding step, image descriptors are locally transformed in a new vector decomposing the initial descriptor on the dictionary. It can be understood as an activation function for the dictionary, activating each of the visual words according to their similarity with the local descriptor.% In the classical BoW representation, the coding function activates only the visual word closest to the descriptor, assigning zero weight to all others. %This is usually called mid-level features which express each descriptor by a subset of visual words.
\item In the pooling step the codes associated with local image features are combined over some image neighbourhood. The codes within each cell are aggregated to create a single feature vector. % Average and maximum pooling are widely used. 
\item Training and classification can be performed on the final feature vectors (usually the concatenation of the signature feature vectors of different cells) by a classifier, e.g. SVM \cite{Cortes1995}.
\end{enumerate}

Also, we introduce some notation used throughout the work. Let an image $I$ be represented by a set of low-level descriptors or local features (e.g. SIFT) $\bm{x_i}$ at $\mathcal{N}$ locations identified with their indices $i=1, \dots, N$.
%,$\bm{\alpha}_i$ in the coded vector (Eq. \ref{eq:coding}). 
%Based in Spatial Pyramid \cite{Lazebnik2006} representation, let $M$ regions of interest be defined on the image, with $\mathcal{N}_m$ denoting the set of indices within the region $m$ (Eq. \ref{eq:pooling}). 
The signature vector $z$ representing the whole image is obtained by sequentially coding (Eq. \ref{eq:coding}) and pooling (Eq. \ref{eq:pooling}) over all descriptors:
\begin{align}
  \bm{\alpha}_i & = f(\bm{x}_i) \;  , i=1, \dots , N  \label{eq:coding}\\
  	\bm{z} & = g(\{\bm{\alpha}_i\}_{i\in \mathcal{N}} ) \label{eq:pooling}
\end{align}
We will denote as $f$ and $g$, the coding and pooling operators respectively and $\bm{\alpha}_i$ the coded vector. 
The classification performance using $\bm{z}$ as the input of a classifier (e.g. SVM \cite{Cortes1995}) depends on the properties and the combination of $f$ and $g$.

In average pooling, the obtained vector $\bm{z}$ is the average over the activations $\bm{\alpha}_i$ of the elements of the image for each component of the coded vector.  
\begin{equation}
\bm{z} = \frac{1}{|\mathcal{N}|}\sum_{i\in\mathcal{N}}\bm{\alpha}_i 
\end{equation}

Whereas in maximum pooling selects the largest value between the activations of the elements of the image for each component of the coded vector. 
\begin{equation}
z_{j} = \underset{i \in \mathcal{N}}{\text{max}} \{ \alpha_{i,j} \} \, ,\text{ for } j = 1, \dots , K
\end{equation}
being $K$ the number of visual words, i.e. the length of the coded vector.
\subsection{Ordered Weighted Pooling}

OWA functions  belong to the class of averaging aggregation functions. They differ to the weighted arithmetic means in the weights, that are associated not with particular inputs, but with the input magnitude. Formally, an OWA operator of dimension $n$ is a mapping $ \phi_w: \mathcal{R}^n\rightarrow \mathcal{R} $ that has an associated collection of weights $\bm{w} =(w_{1},\ldots ,w_{n})$ lying in the unit interval and summing to one. They were introduced by Yager \cite{Yager1988}. We recall the notation of an ordered vector as $(\bm{z}\searrow) = z_1 \geq z_2 \geq \dots \geq z_n$.
\begin{equation}
\phi_w(\bm{z}\searrow)  = \sum_{i=1}^n w_iz_i \,\,\,\,\,\,\, \text{ s.t.} \sum_{i=1}^n w_i = 1 \text{ and }  w_i\ge 0 \text{ for every } i=1,...,n.
\end{equation}  
Obviously the calculation of the value of an OWA function involves sorting the array of values to be aggregated. We can obtain typical aggregation functions \cite{beliakov2016practical}, with specific weights, for example:
\begin{itemize}
\item If $\bm{w} = (0,0, \cdots, 0,1)$,then $\phi_w$ = minimum. 
\item If $\bm{w} = (1,0, \cdots, 0,0)$, then $\phi_w$ = maximum.
\item If $\bm{w} = (\frac{1}{n},\frac{1}{n}, \cdots, \frac{1}{n},\frac{1}{n})$, then $\phi_w$ = arithmetic mean.  
\end{itemize}

Let $\bm{z}^{(I)}$ the final feature vector of an image $I$. This final vector is the aggregation of all of the coded vectors $\bm{\alpha}$ of the dense grid of size $\mathcal{N}$. So, following Eq. (\ref{eq:pooling}), an element $j$ of the vector $\bm{z}^{(I)}$ is calculated with the expression:
\begin{equation}
z^{(I)}_j = \bm{w} \cdot (\bm{\alpha_j} \searrow) 
\end{equation}
%z^{(i)}_K & = w_1\alpha_{K(1)}+ \dots + w_n\alpha_{K(\mathcal{N})} 
\subsection{Learning the weights}
Let's suppose that the classifier used is  a linear kernel support vector machine L2-SVM. The L2-SVM uses the square sum of the slack variables (the regularization  term) which makes the SVM less susceptible to outliers and improve its overall generalization. Next equation shows the cost function $J(\Theta)$ of an L2-SVM, being $\Theta$ the parameters of the model, $\bm{z}^{(i)}$ the final feature vector of an example, $y^{(i)}$ the class of that example, $m$ the number of training examples, $K$ number of features, i.e. the length of the feature vector and $C_1$ the regularization parameter: 
\begin{equation}
\footnotesize
J(\Theta) = \frac{C_1}{m} \sum_{i=1}^{m} max\left(0, 1-\Theta^T\bm{z}^{(i)}y^{(i)}\right)^2 + \frac{1}{2} \sum_{j=1}^K \Theta_j^2
\end{equation}

Due to we want to learn also the weights of the pooling operator $\bm{w}$, if we substitute $\bm{z}$ using Equation (6), the cost function remains as follows:
\begin{equation}
\footnotesize
J(\Theta,w) = \frac{C_1}{m} \sum_{i=1}^{m} max\left(0, 1-\Theta^T(\bm{w}\cdot\bm{\alpha} \searrow)^{(i)}y^{(i)}\right)^2 + \frac{1}{2} \sum_{j=1}^K \Theta_j^2
\end{equation}

Moreover we add a regularization term to smooth distribution of weights for similar activations, such a way we should interpret easily the weights obtained.

\begin{align}
\footnotesize
J(\Theta,w) = & \frac{C_1}{m} \sum_{i=1}^{m} max\left(0, 1-\Theta^T(\bm{w}\cdot\bm{\alpha} \searrow)^{(i)}y^{(i)}\right)^2 + \frac{1}{2} \sum_{j=1}^K \Theta_j^2 + \\ \nonumber 
& + C_2\sum_{i=1}^{\mathcal{N}}\sum_{j=1}^{\mathcal{N}} A_{ij}(w_i-w_j)^2
\end{align}

where $C_2$ is th regularization parameter and $A$ is a matrix with elements $A_{ij} = 1$ if $j=i+1$ and $A_{ij} = 0$ otherwise.  However, the pooling weights must satisfy the restrictions  $\sum_{i=1}^{\mathcal{N}} w_i = 1$, and $\forall w_i\geq 0$ therefore the training step is a constrained optimization problem. Using the method of Lagrange multipliers to convert it into an unconstrained problem:

\begin{equation}
\mathcal{L}(\Theta, w, \lambda, \mu) = J(\Theta, w) + \lambda\left(\left(\sum_{i=1}^{\mathcal{N}} w_i\right) - 1\right) + \sum_{i=1}^{\mathcal{N}} - w_i*\mu_i
\end{equation}

The following equations are the partial derivatives of the parameters $\Theta$, $\bm{w}$ and the Lagrange multipliers $\lambda$ and $\mu_i$, which can be used in any gradient descent based optimization algorithm:
\begin{equation}
\footnotesize
\frac{\partial \mathcal{L}(\Theta, w, \lambda, \mu)}{\partial \Theta_j} = \frac{-2C_1}{m} \sum_{i=1}^m max(0, 1 - \Theta^T (\bm{w}\cdot\bm{\alpha} \searrow)^{(i)}y^{(i)})(w\cdot\bm{\alpha} \searrow)^{(i)}y^{(i)} + \Theta_j
\end{equation}

\begin{align}
\footnotesize
\frac{\partial \mathcal{L}(\Theta, w, \lambda, \mu)}{\partial w_j} & =   \frac{-2C_1}{m} \sum_{i=1}^m max(0, 1 - \Theta^T (\bm{w}\cdot\bm{\alpha} \searrow)^{(i)}y^{(i)})\Theta^T(w_j\cdot \bm{\alpha} \searrow)^{(i)}y^{(i)}  +  \nonumber \\ & +   2C_2 \sum_{i=1}^{\mathcal{N}}\sum_{j=1}^{\mathcal{N}} A_{ij}(w_i-w_j) + \lambda - \mu_j 
\end{align}

\begin{equation}
\footnotesize
\frac{\partial \mathcal{L}(\Theta, w, \lambda, \mu)}{\partial \lambda} = \left( \sum_{i=1}^{\mathcal{N}} w_i \right) - 1
\end{equation}

\begin{equation}
\footnotesize
\frac{\partial \mathcal{L}(\Theta, w, \lambda, \mu)}{\partial \mu_j} = - w_j 
\end{equation}

Instead of learning all of the parameters ($\Theta$ and $w$) in the training process, we can simplify and learn the parameters in an iterative way, first learning the parameters $\Theta$ of the classifier, then find the optimal values of the OWA $\bm{w}$. Then using the OWA obtained, retrain the parameters $\Theta$ of the classifier (being the initial $\Theta$, the one obtained in the first optimization), next, retrain the parameters of the OWA and so on, until convergence, i.e. the parameters do not change. We have used this methodology due to it avoids falling in saddle points and achieves better optimization than training all of the parameters ($\Theta$ and $w$) together from scratch. 

\subsection{First Experiment}
\begin{figure}
\begin{center}
\includegraphics[width=.7\textwidth]{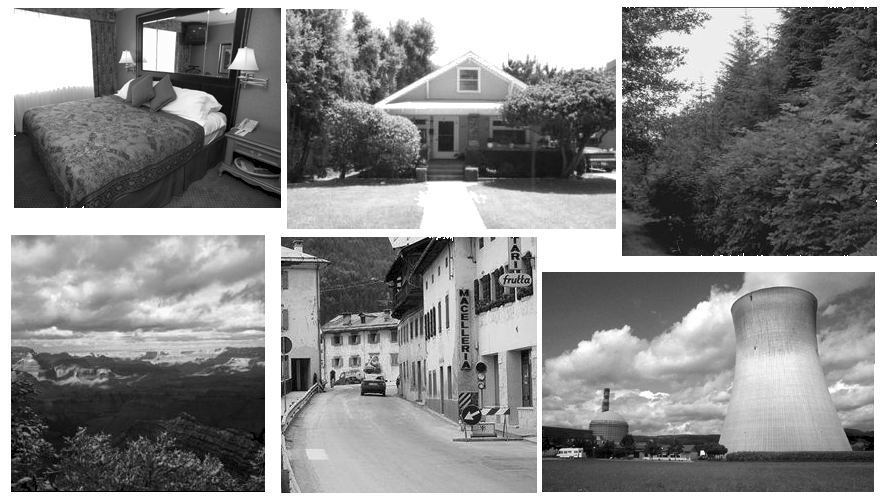}
\caption{Sample images from 15-Scenes dataset. }
\label{fig:15ScenesSample}
\end{center}
\end{figure}

In this experiment we evaluate the performance of OWA-pooling by means of the classification accuracy in the 15-Scenes dataset (see Figure \ref{fig:15ScenesSample}),on a BoW setting \cite{Lazebnik2006}. Low-level descriptors $x_i$ are 128-dimensional SIFT descriptors \cite{Lowe2004} of 32 $\times$ 32 patches. The descriptors are extracted on a dense grid every 32 pixels, such that the image is divided in 64 cells.  The dictionary $\bm{D}$ is calculated from $200.000$ random descriptors. The vocabulary sizes tested are $[2^4, 2^5, .., 2^{9}]$. The coding step is carried out by two different methods, triangle assignment \cite{Coates2011} and sparse coding \cite{5206757}. The same dictionary was used in both coding methods. Following the usual procedure, we use 150 training images and the rest for testing on the 15-Scene dataset. Experiments are conducted over 10 random splits of the data, and we report the test mean accuracy. The value of the regularization parameters $C_1$ and $C_2$ are selected by cross-validation within the training set.

In this first experiment we  aggregate the feature vectors of the whole image in the pooling step, without creating regions (like the spatial pyramid method), so that the overlap between regions does not interfere in the results and the performance of the pooling operators is clearly reflected in the results.

\begin{figure}
\begin{tabular}{ccc}
\includegraphics[width=0.3\textwidth]{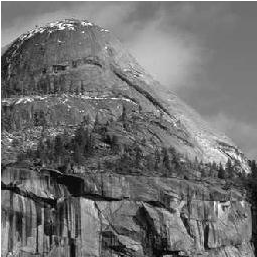} & 
\includegraphics[width=0.3\textwidth]{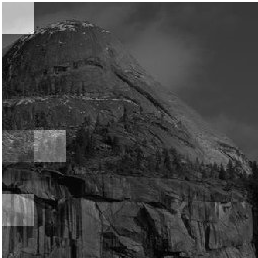}&
\includegraphics[width=0.3\textwidth]{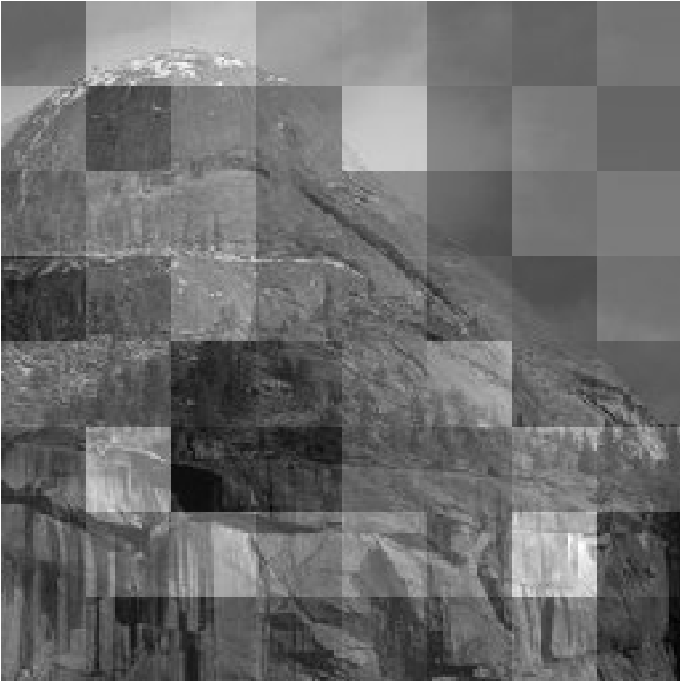} \\
(a) & (b) & (c)
\end{tabular}
\caption{Contribution of each cell in the final vector representation. The importance of each cell is superimposed in the image, transparent cells are more important and darker cells are less important. (a) Original image, (b) using MAX pooling and (c) using OWA-pooling. }
\label{fig:OWArep}
\end{figure}

Our hypothesis is that the pooling weights learned on the iterative process will be adapted to the data and the features with high activation but low representativeness of the object of the image should be filtered. The results depicted in Table \ref{table:E1}, shows that the proposed method improves the results over maximum and average pooling. Especially for features coded with spare coding, we obtain more than a 10\% increase in the performance. To analyse the relation between the feature values   and their representativeness, in Figure \ref{fig:OWArep} we represent the contribution of each one of the 64 cells on the sum of  $\bm{w}\cdot(\alpha \searrow)$ (i.e. the value in the final vector) in an image that is misclassified by the maximum pooling. We can see that using the maximum, few cells contribute to the final representation, but the OWA pooling distribute better the contribution of different parts of the image (in the case of average pooling all of the cells have the same importance).

\begin{table}[!h]
\caption{Accuracies in 15-Scenes dataset with different codings, dictionary sizes and pooling operations.}
\begin{center}
\tiny
\begin{tabular}{|c|c|c|c|c|c||c|c|c|c|c|}
\hline
Coding& \multicolumn{5}{|c||}{Triangle assignment (\% accuracy)} & \multicolumn{5}{|c|}{Sparse (\% accuracy)}\\
\hline
Dic.Size & 32 & 64 & 128 & 256 & 512 & 32 & 64 & 128 & 256 & 512 \\ \hline
MAX & 43.89 & 48.94 & 55.38 & 60.23 & 63.09 &48.7 & 55.91 & 61.78 & 64.82 & 68.76  \\ \hline
MEAN & 39.41 & 48.02 & 48.17 & 48.56 & 49.43 & 54.15 & 61.06 &  63.06 & 64.21 &  64.39\\ \hline
OWA & \textbf{45.99}  & \textbf{49.75} & \textbf{55.57} & \textbf{60.40} & \textbf{64.74} & \textbf{58.18}  & \textbf{65.05} & \textbf{71.05} & \textbf{74.89} & \textbf{80.26} \\ \hline
\end{tabular}
\label{table:E1}
\end{center}
\end{table}

\section{Learning Ordered weighted pooling in Convolutional Neural Networks}\label{sec:CNN}
In this section we describe how we can use an OWA-pooling in a CNN. Let's suppose that we have an OWA-pooling after one convolutional layer and we want to learn its parameters $\bm{w}$ at the same time that we train the network.  The pooling parameters could be learned similar to the convolutional filters, because the calculation of the gradient $\frac{\partial J}{\partial w_i}$  is similar to that of a convolution parameter, taking into account that it is necessary to sort the values of the activation layer. However,  the pooling parameters must satisfy the restrictions  $\sum_{i=1}^{\mathcal{N}} w_i = 1$, and $\forall w_i\geq 0$, so we add three terms to the cost function:
\begin{eqnarray}
&J = J_{CE} + C_1  \sum_{i=1}^N max \left\lbrace 0, - w_i \right\rbrace +  \nonumber \\ +& C_2 \left ( \left (\sum_{i=1}^{\mathcal{N}} w_i\right) - 1 \right)^2 +  C_3 \left ( \sum_{i=1}^{\mathcal{N}-1} \left (w_i - w_{i+1} \right)^2  \right) \label{eq:J}
\end{eqnarray}

Where $J_{CE}$ is the cost function of the CNN (for example the cross correlation) and $C_1$, $C_2$ and $C_3$ are regularization parameters. The first term trigger the values to be greater than zero, the second term penalizes if the sum of weights is not one and the third term penalizes the differences between the consecutive weights. Therefore, the calculation of the gradient to be used in the optimization algorithm is:
\begin{equation}
\frac{\partial J}{\partial w_i}  = \frac{\partial J_{CE}}{\partial w_i} - C_1  + 2C_2\left( \left(\sum_{i=1}^{\mathcal{N}} w_i\right) - 1 \right)w_i  + 2C_3(w_i-w_{i+1})  
\end{equation}
If $w_i \geq 0$ the term $ C_1 $ disappears.

%There are several options if we use OWA-pooling in a CNN. First, to learn a set of weights for the network, i.e. to use the same OWA in every pooling operation, second, to learn a set of weights per layer, third to learn a set of weights per channel of each layer and finally to learn a set of weights for each region of each channel of each layer.

We are going to compare the performance of the OWA-pooling in two different scenarios, in the first experiment, we are going to learn the weights of an OWA-pooling in a small CNN. In the second experiment, we will test the performance of the OWA pooling used as a global pooling operation in deeper networks such as VGG and mobilnet.

\subsection{First experiment}
In this experiment we want to check the accuracy of already known and validated CNN when we replace the original  pooling operator with our proposed OWA-pooling. We also want to check if the weights converge towards pooling operators similar to the original operators ($ (1,0,0, \cdots, 0) $ for the maximum or $ (\frac{1}{n} , \frac{1}{n}, \cdots, \frac{1}{n}) $ for the mean). We are going to use the CNN proposed in \cite{NiN} which is known as Network in Network (NiN). Its architecture is shown in table \ref{table:NINArchitecture}, where all convolutions are followed by activations type Relu.

\begin{table}[!h]
\caption{NiN architecture}
\begin{center}
\tiny
\begin{tabular}{|c|c|c|}
\hline
Input & & Filters  channels  \\
\hline \hline
32x32 & &5x5, 192 \\ \hline
32x32 & &1x1, 160 \\ \hline
32x32 & &1x1, 96 \\ \hline
32x32 & pool1 &3x3 Max {\it pooling}, stride 2  \\ \hline
16x16 & &dropout 0.5 \\ \hline
16x16 & &5x5, 192 \\ \hline
16x16 & &1x1, 192 \\ \hline
16x16 & &1x1, 192 \\ \hline
32x32 & pool2 &3x3 Ave {\it pooling} \\ \hline
8x8 & &dropout 0.5 \\ \hline
8x8 & &5x5, 192 \\ \hline
8x8 & &1x1, 192 \\ \hline
8x8 & &1x1, 10 o 100 \\ \hline
8x8 & pool3 &8x8 Ave {\it pooling} \\ \hline
10 o 100 & & Softmax \\ \hline
\end{tabular}
\label{table:NINArchitecture}
\end{center}
\end{table}
%
%\begin{table}[!h]
%\caption{Parámetros de la red VGG}
%\begin{center}
%\tiny
%\begin{tabular}{c|c|c}
%\hline
%Input  &  & Filtros,  canales  \\
%\hline \hline
%28x28 & & 3x3, 64 \\ \hline
%28x28 & &3x3, 64\\ \hline
%28x28 & pool1 &2x2 Max {\it pooling}, stride 2\\ \hline
%14x14 & &3x3, 128 \\ \hline
%14x14 & &3x3, 128 \\ \hline
%14x14 & pool2 &2x2 Max {\it pooling}, stride 2  \\ \hline
%7x7 & &3x3, 256 \\ \hline
%7x7 & &3x3, 256 \\ \hline
%7x7 & pool3 &2x2 Max {\it pooling}, stride 2\\ \hline
%4x4 & &3x3, 512 \\ \hline
%4x4 & &3x3, 512 \\ \hline
%4x4 & pool4 &2x2 Max {\it pooling}, stride 2 \\ \hline
%2x2 & &3x3, 512 \\ \hline
%2x2 & &3x3, 512 \\ \hline
%2x2 & pool5 &2x2 Max {\it pooling}, stride 2 \\ \hline
%- & &Flatten \\ \hline
%- & &2048 \\ \hline
%- & &512 \\ \hline
%- & &10 \\ \hline
%- & & Softmax \\ \hline
%\end{tabular}
%\label{table:VGG Architecture}
%\end{center}
%\end{table}

We have trained the network with  CIFAR-10 and CIFAR-100 \cite{Krizhevsky09} datasets, which contains 50,000 color images for training and 10000 for test. The images have a resolution of 32x32 pixels and there 10 and 100 categories respectively. 
In  table \ref{table:NIN1} we compare different pooling configurations:
\begin{itemize}
\item Orig: is the original pooling operations as shown in table \ref{table:NINArchitecture}.
\item MAX: Maximum pooling in the three layers.
\item AVE: Average pooling in the three layers.
\item OWAL: learn OWA weights for each layer.
\item OWALnr: learn OWA weights for each layer without restrictions.
\item OWALC: learn OWA weights for each channel of each layer.
\item OWALCnr: learn OWA weights for each channel of each layer without restrictions.
\end{itemize}

The results obtained verify that when OWA-pooling is used, a similar or better accuracy is reached compared with the original. Moreover, when the weights are trained without restrictions, the accuracy is slightly higher. It is also verified that fitting a pooling operator for each channel of each layer is not necessary in this case, in fact, the accuracy of OWALC is lower than OWAL.

\begin{table}[!h]
\caption{Accuracies in CIFAR-10 and CIFAR-100 with different pooling operations.}
\begin{center}
\footnotesize
\begin{tabular}{|l||r|r|}
\hline
Pooling & CIFAR-10 & CIFAR-100 \\
 & (\% error) & (\% error)\\
\hline \hline

Orig & 9.76 & 41.30\\ \hline
MAX & 11.24 & 45.11\\ \hline
AVE & 9.50 & 41.25\\ \hline
OWAL & 9.66 & 40.67\\ \hline
OWALnr & \textbf{9.42} & \textbf{40.44}\\ \hline
OWALC & 9.68 & 41.91\\ \hline
OWALCnr & 9.64 & 41.33\\ \hline
%Orig & 90.24 & 58.70\\ \hline
%MAX & 88.76 & 54.89\\ \hline
%AVE & 90.50 & 58.75\\ \hline
%OWAL & 90.34 & 59.33\\ \hline
%OWALnorestr & \textbf{90.58} & \textbf{59.66}\\ \hline
%OWALC & 90.32 & 58.09\\ \hline
%OWALCnorestr & 90.36 & 58.77\\ \hline
%MPO1ft & 90.39 & 59.42\\ \hline
%MPO1pl-ft & 90.46 & 59.62\\ \hline
\end{tabular}
\label{table:NIN1}
\end{center}
\end{table}

%\begin{table}[!h]
%\caption{Resultados experimentales de la red VGG.}
%\begin{center}
%\begin{tabular}{l||r}
%\hline
%VGG & FMNIST \\
%\hline \hline
%Max & \textbf{94.31} \\ \hline
%Mean & 92.90 \\ \hline
%MPO1 & 93.57 \\ \hline
%MPO2 & 92.11 \\ \hline
%MPO1ft & 92.87 \\ \hline
%MPO1pl & 94.25 \\ \hline
%MPO2pl & 92.91 \\ \hline
%MPO1pl-ft & 92.770 \\ \hline
%\end{tabular}
%\label{table:VGG1}
%\end{center}
%\end{table}

\subsubsection{OWA pooling and robustness}
\begin{figure}[!htbp]
\centering
\begin{tabular}{cc}
\includegraphics[width=0.45\textwidth]{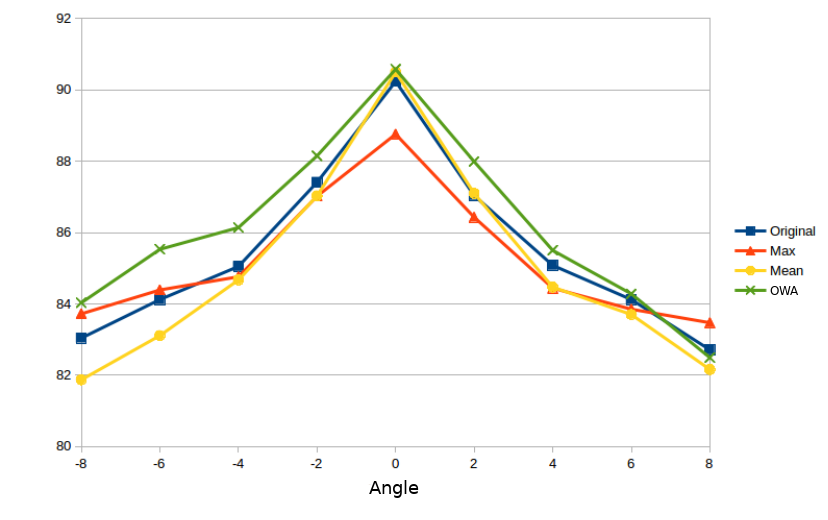} &
\includegraphics[width=0.45\textwidth]{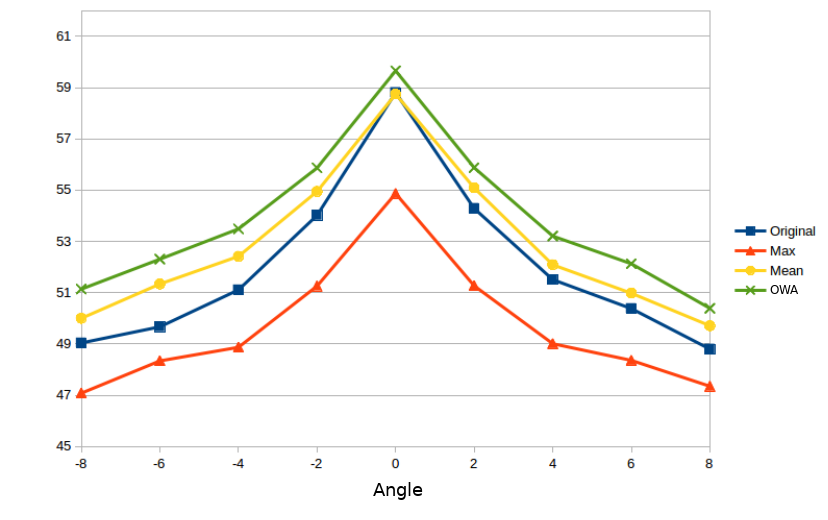} \\
(a) & (b)
\end{tabular}
\caption{Robustness to rotation. Vertical axis of the graphs represents the accuracy and the horizontal axis the angle that the images are rotated. (a) Results for NiN trained in CIFAR-10 and (b) Results for NiN trained in CIFAR-100.}
\label{fig:tr2}
\end{figure}

Adding more trainable parameters to the network usually means networks proned to overfitting and therefore with less generalization capacity. To check this fact we have carried out a robustness test \cite{pmlr-v51-lee16a}. In this test, the test images are rotated between -8 and 8 degrees and we compute the accuracy with the trained networks (Figure  \ref{fig:tr2}). We verify that in the case of NiN, the model with OWAL is the most robust to changes. Therefore, to weight the activations taking into account only their value, causes the representative feature of the image to spread through the network to the final representation and improve the classification.

\subsubsection{OWA pooling and regularization}
In the previous section we have proved that adding more trainable parameters in OWA not means more overfitting, being OWA and NIN more robust to changes than other pooling methods. In that way, in this section we combine OWA with other common regularization techniques to prove if both methods are complementary leading to an improved result. We have tested two different recent regularization approaches in CNNs:(1) data augmentation by means of random erasing \cite{zhong2020random} and (2) an structured dropout called DropBlock \cite{NIPS2018_8271}. In training, random erasing randomly selects a rectangle region in an image and erases its pixels with random values. We have trained the NiN model (both, the original architecture and using OWA pooling in every pooling layer) in the CIFAR-10 dataset with random erasing. We found that this regularization does not have a positive effect on the results, actually the accuracy in both models, with and without random erasing was almost the same of Table \ref{table:NIN1}. This may be because the NiN network is not a deep network compared to ResNet or VGG, where other authors have proved that random erasing increases accuracy in CIFAR-10.

%In table \ref{} we show the percentage of error of the original training of the NiN model in Cifar-10 and the error after training with random erasing, with different erasing probability $p$ and different area of erasing region $s_h$ \footnote{The rest of the hyper parameters are the ones proposed in the original work \cite{zhong2020random} $s_l=0.02$, $r_1=0.3$ and  $r_2=1/0.3$.}.

%\begin{table}
%\centering
%\begin{tabular}{ccccc}
% & NiN & Max & Ave & OWA \\ \hline
%
%Original &9.76	&11.24& 9.5&	9.42\\ \hline \hline
%$p=0.5, s_h=0.4, $	& 9.99	&11.78&	9.32&	9.39\\ \hline
%					
%$p=0.5,  s_h=0.3$	&9.67	&11.76	&9.4&	9.34\\\hline
%					
%$	p=0.5,  s_h=0.5$	&10.09	&11.3&	9.99&	9.51\\\hline
%					
%$	p=0.2,  s_h=0.4$	&10.11	&11.24&	9.86	&9.99\\\hline
%
%$	p=0.2, s_h=0.3$&	10.07&	10.97&	9.51&9.81\\\hline
%					
%$	p=0.2, s_h=0.5$	&10.0&	11.21&	9.43&	9.65\\\hline
%
%\end{tabular}
%\caption{}
%\end{table}

However, the structured dropout DropBlock \cite{NIPS2018_8271}, where units in a contiguous region of a feature map are dropped together, improves the accuracy in our experiments. We have tested different block sizes and probabilities and the results shows that the model trained with DropBlock increases accuracy. The error of the best OWA model decreases from 9.42 to 8.26 which is a very important improvement (Table \ref{tab:dropblock}).
Therefore OWA pooling is complementary with DropBlock and can be used to increase the accuracy. Even, the largest improvement is obtained when the NiN network pooling operators are OWAs.

\begin{table}
\centering
\begin{tabular}{ccccc}
 & NiN & Max & Ave & OWA \\ \hline
Original &9.76	&11.24& 9.5&	\textbf{9.42}\\ \hline \hline
$s_1=4, p_1=0.8, s_2=2, p_2=0.8$& 9.09&	11.26&	9.3	&9\\ \hline 
$s_1=5, p_1=0.8 , s_2=3, p_2=0.8$& 8.92&	10.32&	8.5&	8.31\\ \hline 
$s_1=5, p_1=0.5 , s_2=3, p_2=0.5$ &9.85&	12.71&	9.42&	8.89\\ \hline 
$s_1=7, p_1=0.8 , s_2=5, p_2=0.8 $&8.71&	10.44&	8.63&	\textbf{8.26}\\ \hline 
$s_1=7, p_1=0.5 , s_2=5, p_2=0.5$& 9.93&	14.3&	9.35&	9.14\\ \hline 
\end{tabular}
\caption{Comparison of error percentage in CIFAR-10 of the NiN network trained without regularization (Original) and with DropBlock in different configurations. MAX, Ave and PWA  where all of the pooling operators are maximum,  the average  and OWA respectively. $s_1$ and $s_2$ are the size of the blocks in layers 1 and 2 respectively and $p$ is the keep probability.} \label{tab:dropblock}
\end{table}

\subsubsection{Weight analysis}
\begin{figure}[htbp]
\centering
\begin{tabular}{cc}
\includegraphics[width=0.45\textwidth]{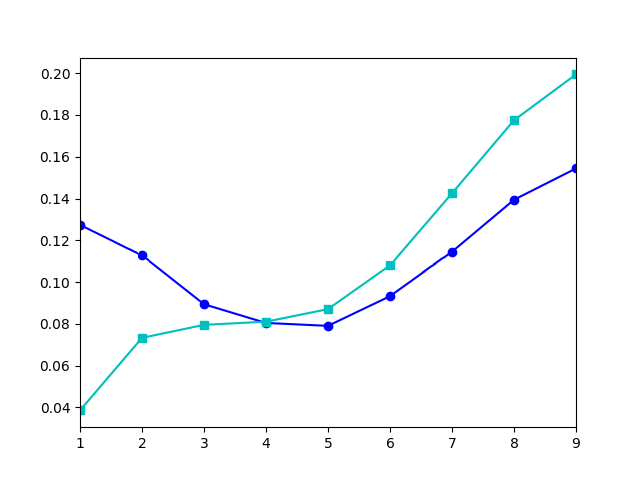} &
\includegraphics[width=0.45\textwidth]{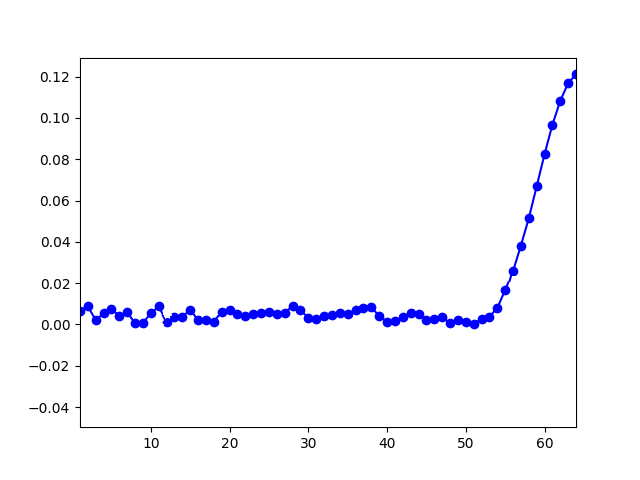}\\
(a) & (b)
\end{tabular}
\caption{(a) 9 learned weights of the OWA-pooling for layers 1 and 2 of NiN trained with CIFAR-10. (b) 64 learned weights of tthe OWA-pooling for layer 3 of NiN trained with CIFAR-10.}
\label{fig:NiNweights}
\end{figure}

In the Figure \ref{fig:NiNweights} are depicted the weights of the three pooling layers learned for NiN network in the OWAL case for CIFAR-10 (weights for CIFAR-100 are similar). In both figures the weights are depicted in order, i.e. the weight in position 1 will multiply the largest value of the vector, the second weight with the second largest and so on. Figure 4(a) represent the nine weights for first and second pooling layer, for both layers weights are around the average  but both giving more importance to the smaller activations. Figure 4(b) shows the sixty-four weights of the third layer which also boost lower activations. 
 
\subsubsection{Performance analysis} The biggest problem with OWA pooling is the need of sorting the activation values. The introduction of the magnitude order into the training algorithm, makes computing times grow. In the table \ref{table:Time} we show the increase of time of the different models with respect to the original. The time cost of the OWAL model is not too large and it is probably faster than test the combinations of maximum and average pooling operators for each layer of the network (for small networks). Therefore the main conclusion of this experiment is that, OWA pooling can be used in real cases in which we do not know the correct architecture and we decided to introduce in the training step the pooling operator learning.

\begin{table}[!h]
\caption{Comparison of training times with respect to to the original NiN architecture.}
\begin{center}
\small
\begin{tabular}{|l||r|r|r|r|r}
\hline
NiN & Orig & MAX & AVE & OWA \\
\hline \hline
Time & 1 & 1.02x & 0.95x & 4.06x \\ \hline
%VGG & Orig & MAX & AVE & MPO1 & MPO1ft\\
%\hline 
%Tiempo & 1 & 1 & 0.83x & 7.26x & 2.11x\\ \hline\hline
\end{tabular}
\label{table:Time}
\end{center}
\end{table}

\subsection{Second experiment. Global pooling layer}

This second experiment is motivated by two factors. Firstly, pooling operations in intermediate layers of neural networks usually aggregate only a few values, (usually 2x2 or 3x3 pool size), compared with our first BoW experiment where 64 values were aggregated. So we wonder if this operation is more important when bigger regions are aggregated like the global pooling. Second, as we discuss in the previous section, OWA-pooling increases the execution time of the pooling layers, this problem is bigger as deeper is our neural network and as larger are the input dimensions of our image. For example, for an input image of 224x224x3 pixels that after initial convolutions have a size of 222 width x 222 height x 32 filters a 2 x 2 pooling size will produce 394272 pooling operations. Taking into account this two aspects, we decided to study the performance of applying our OWA-pooling only as global pooling layer. This global pooling layer is common among different well known architectures.

We study the performance in two well-known neural network architectures, VGG13 and MobileNet. VGG architecture do not contain a global pooling layer like MobileNet, so in our experiments we use the original architecture till the pool5 layer, where we concatenate a global pooling layer followed by a softmax.

Two datasets were used, 15-Scenes and Caltech-UCSD Birds (CUB200). CUB200 is a dataset with 6033 bird images classified in 200 different species. For 15-Scenes experiments we set as image size 256x256 pixels, using 150 images per class for training purposes. In Cub200 experiments we rescale images to 224x224 pixels and we split the dataset using 80\% of the images for training and the other 20\% for validation and test purposes. We made random data augmentation in both datasets by horizontal flips, rotations, translations and zooms. The parameters used in 15 scenes were: 10 degrees of rotations, 15\% translations,  and 10\% zoom. In CUB200 experiments were: 15 degrees of rotations, 15\% translations, 15\% zoom and 15\% shear. All of the experiments in this two datasets began with pre-initialized Imagenet weights for both architectures. Then, the model is trained a few epochs with different aggregations in the global pooling layer. The number of epochs that is trained each model depends on its own convergence. CUB200 VGG was trained 300 epochs and CUB200 MobileNet 150 epochs. For 15-Scenes, the net VGG was trained for 25 epochs and MobileNet 50 epochs.

In Table \ref{table:AccGlobal} we show the results obtained for different global pooling operators: AVE (average mean), MAX (maximum), OWA, OWAco (OWA-pooling constrains are implemented by means of the constrains module from keras instead of the regularization formulation of Eq. (\ref{eq:J})), OWAnr (uncostraied weights) and OWAC (a OWA-pooling operator is learned per channel).

\begin{table}[!h]
\caption{Classification error in CUB and 15-Scenes with different global pooling operations}
\begin{center}
\tiny
\begin{tabular}{|l|l||c|c|}

\hline

\hline
Model & Global Pooling & CUB (300 epochs)	& 15-Scenes (25 epochs)\\
&&	(\% error)	& (\% error)\\
\hline \hline

VGG & Orig (+ 4million param.)	& 27.92	& 8.92\\
\hline \hline

VGG &  AVE	& 28.29	& 14.94\\
\hline
VGG &  MAX	& 25.06	& 11.61\\
\hline
VGG &  OWA	& \textbf{23.2}	& \textbf{10.76}\\
\hline
VGG &  OWAco & 	23.78 & 	10.85\\
\hline
VGG &  OWAnr	& 24.48	& 10.88\\
\hline
VGG &  OWAC & 23.94	&12.21\\
\hline
VGG & OWACco & 23.49 &	11.32\\
\hline
VGG & OWACnr &	24.28	&10.92\\

\hline \hline
	
&	&CUB (150 epochs)&	15Scenes (50 epochs)\\
&&	(\% error)	& (\% error)\\
\hline
MobileNet &  AVE	&22.29	&6\\
\hline
MobileNet &  MAX	&25.77	&7.82\\
\hline
MobileNet &  OWA &	23.94	&6.42\\
\hline
MobileNet & OWAco &	23.69	&6.22\\
\hline
MobileNet &  OWAnr	&26.1	&6.53\\
\hline
MobileNet & OWAC &\textbf{22.43}	&5.88\\
\hline
MobileNet &  OWACco &23.62	&\textbf{5.75}\\
\hline
MobileNet &  OWACnr &23.94 & 6.42\\
\hline
\end{tabular}
\label{table:AccGlobal}
\end{center}
\end{table}

Results using OWA-pooling for VGG architecture are better compared with average or maximum. It is important to take in account that the original version of VGG for the 15-Scenes experiments is better than the rest of values, this accuracy should be seen as a reference, because using the global pooling instead of the fully connected layers means that our proposed test architecture has 4 million less parameters. Also it is remarkable to see that for the CUB200 experiment the better experiment using OWA-pooling outperforms the original model even with significantly fewer parameters.

For MobileNet, in the CUB200 dataset, the best global pooling is the average, which is the pooling operator of the original architecture, but the best OWA reaches almost similar error. However in  15-Scenes, the best result is obtained learning a OWA-pooling for every channel. Summing up, we can conclude again that our proposed aggregation technique produce better results than the classical average or maximum aggregation, and in the worst case results are similar. This take sense if we consider that for some problems maximum pooling and average pooling can be the best option, and this aggregation functions are particular cases of OWA pooling that can be learned during the training process.

\subsubsection{Weights analysis}
\begin{figure}[htbp]
\centering
\begin{tabular}{cc}
\includegraphics[width=0.42\textwidth]{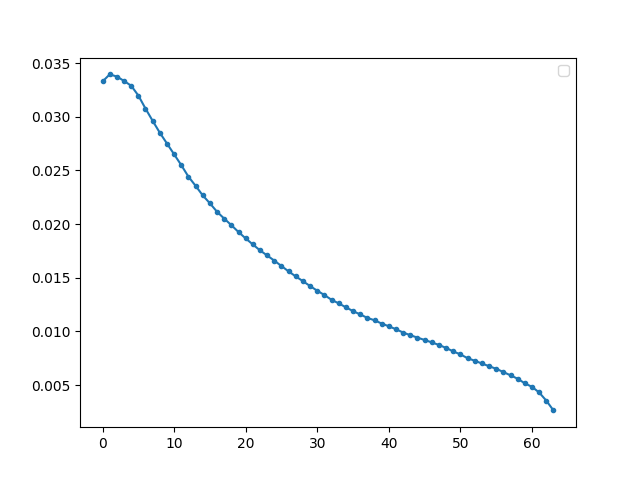} &
\includegraphics[width=0.52\textwidth]{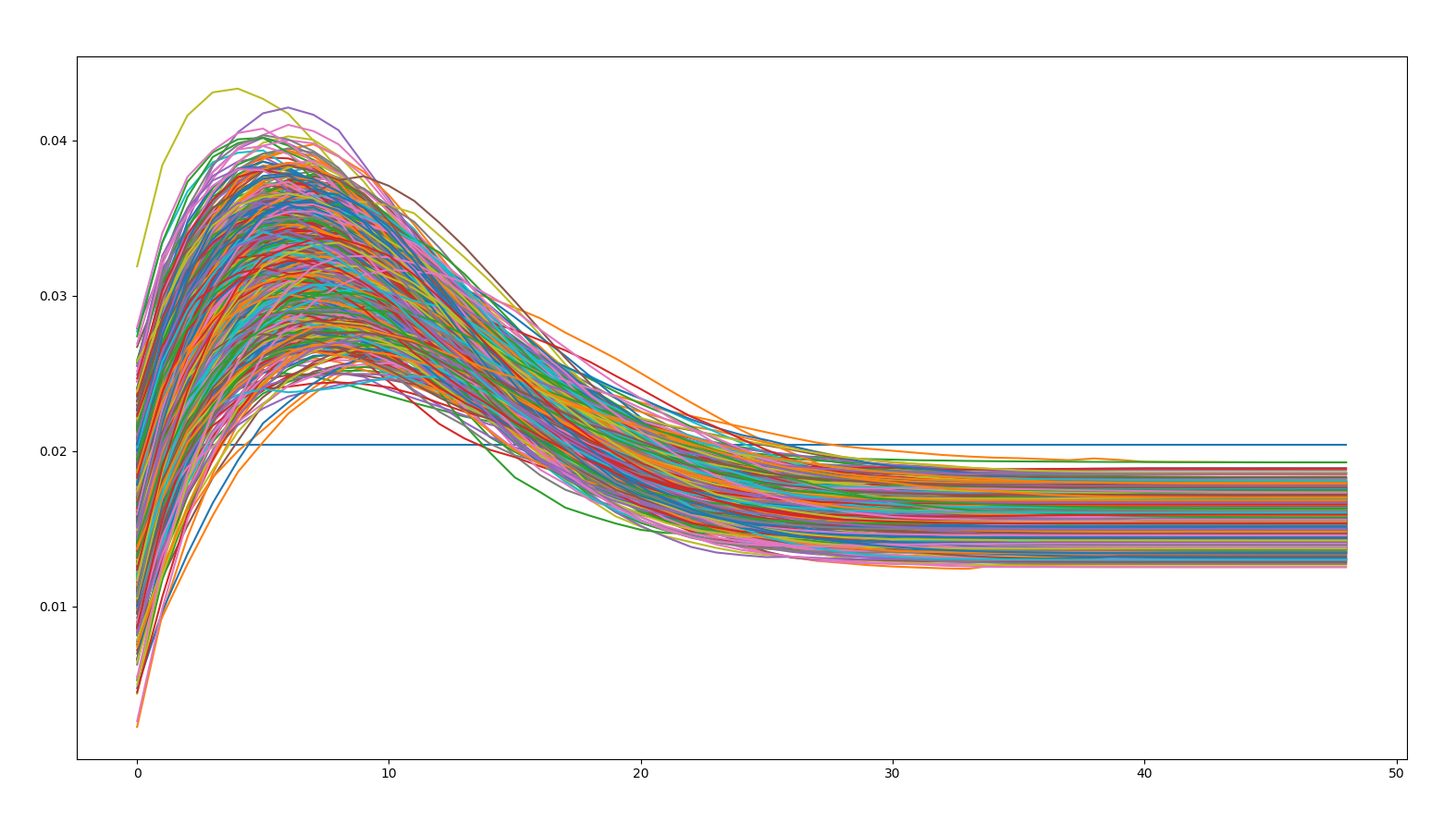}\\
(a) & (b)
\end{tabular}
\caption{(a) 64 learned weights of the OWA-pooling for the global pooling layer using VGG model and 15-Scenes dataset.(b) 64 learned weights of the OWA-pooling for the global pooling layerfor every chanel (OWAC)  using MobileNet model with CUB200 dataset.}
\label{fig:SecondExperimentNweights}
\end{figure}

In Figure \ref{fig:SecondExperimentNweights} are shown the weights of two different cases of this second experiment. In the left image are shown the 64 weights using the VGG model with the 15-Scenes dataset (OWA). In the right image are represented the 49 weights for each pooling function learned in OWAC using MobileNet model for the CUB200 dataset. The mobilenet architecture used is 1024 dimensions depth in the global pooling operation, so in this image are represented these 1024 learned weight vectors.  In both cases, the resultant aggregation functions penalize the highest value but give more weights to he highest values. This result verifies the conclusions obtained by Boureau et. al. \cite{Boureau2010}.

\section{Conclusions}
\label{sec:conc}
In this paper we propose a pooling method based on ordered weighted averaging operators named OWA-pooling. We have studied empirically the performance of OWA-pooling compared to common aggregation methods applied to image classification problems for BoW method and for CNNs, in four different datasets. In all of the experiments,  the OWA-pooling learned obtains an accuracy equal or better than the maximum or average pooling. We can determine that OWA-pooling is a self-configurable aggregation, and can be used to avoid the choice of a pooling function in the designing process of a recognition architecture. 

\section*{Acknowledgment}
This work is partially supported by the research services of Universidad P\'ublica de Navarra and by the project TIN2016-77356-P (AEI/FEDER, UE). We also would like to thank Nvidia for providing equipment.

\bibliography{OWApooling}

\begin{thebibliography}{10}
\expandafter\ifx\csname url\endcsname\relax
  \def\url#1{\texttt{#1}}\fi
\expandafter\ifx\csname urlprefix\endcsname\relax\def\urlprefix{URL }\fi
\expandafter\ifx\csname href\endcsname\relax
  \def\href#1#2{#2} \def\path#1{#1}\fi

\bibitem{726791}
Y.~Lecun, L.~Bottou, Y.~Bengio, P.~Haffner, Gradient-based learning applied to
  document recognition, Proceedings of the IEEE 86~(11) (1998) 2278--2324.
\newblock \href {http://dx.doi.org/10.1109/5.726791}
  {\path{doi:10.1109/5.726791}}.

\bibitem{Lazebnik2006}
S.~Lazebnik, C.~Schmid, J.~Ponce, {Beyond bags of features: Spatial pyramid
  matching for recognizing natural scene categories}, Proceedings of the IEEE
  Computer Society Conference on Computer Vision and Pattern Recognition 2
  (2006) 2169--2178.
\newblock \href {http://arxiv.org/abs/9411012} {\path{arXiv:9411012}}, \href
  {http://dx.doi.org/10.1109/CVPR.2006.68} {\path{doi:10.1109/CVPR.2006.68}}.

\bibitem{AlexNet}
A.~Krizhevsky, I.~Sutskever, G.~E. Hinton,
  \href{http://dl.acm.org/citation.cfm?id=2999134.2999257}{Imagenet
  classification with deep convolutional neural networks}, in: Proceedings of
  the 25th International Conference on Neural Information Processing Systems -
  Volume 1, NIPS'12, Curran Associates Inc., USA, 2012, pp. 1097--1105.
\newline\urlprefix\url{http://dl.acm.org/citation.cfm?id=2999134.2999257}

\bibitem{VGG}
K.~Simonyan, A.~Zisserman, \href{http://arxiv.org/abs/1409.1556}{Very deep
  convolutional networks for large-scale image recognition}, CoRR
  abs/1409.1556.
\newblock \href {http://arxiv.org/abs/1409.1556} {\path{arXiv:1409.1556}}.
\newline\urlprefix\url{http://arxiv.org/abs/1409.1556}

\bibitem{NiN}
M.~Lin, Q.~Chen, S.~Yan, \href{http://arxiv.org/abs/1312.4400}{Network in
  network}, CoRR abs/1312.4400.
\newblock \href {http://arxiv.org/abs/1312.4400} {\path{arXiv:1312.4400}}.
\newline\urlprefix\url{http://arxiv.org/abs/1312.4400}

\bibitem{GoogleNet}
C.~Szegedy, W.~Liu, Y.~Jia, P.~Sermanet, S.~Reed, D.~Anguelov, D.~Erhan,
  V.~Vanhoucke, A.~Rabinovich, Going deeper with convolutions, in: 2015 IEEE
  Conference on Computer Vision and Pattern Recognition (CVPR), 2015, pp. 1--9.
\newblock \href {http://dx.doi.org/10.1109/CVPR.2015.7298594}
  {\path{doi:10.1109/CVPR.2015.7298594}}.

\bibitem{Boureau2010}
Y.~L. Boureau, F.~Bach, Y.~LeCun, J.~Ponce, {Learning mid-level features for
  recognition}, Proceedings of the IEEE Computer Society Conference on Computer
  Vision and Pattern Recognition (2010) 2559--2566\href
  {http://dx.doi.org/10.1109/CVPR.2010.5539963}
  {\path{doi:10.1109/CVPR.2010.5539963}}.

\bibitem{Bengio2012}
Y.~Bengio, \href{https://doi.org/10.1007/978-3-642-35289-8_26}{Practical
  Recommendations for Gradient-Based Training of Deep Architectures}, Springer
  Berlin Heidelberg, Berlin, Heidelberg, 2012, pp. 437--478.
\newblock \href {http://dx.doi.org/10.1007/978-3-642-35289-8_26}
  {\path{doi:10.1007/978-3-642-35289-8_26}}.
\newline\urlprefix\url{https://doi.org/10.1007/978-3-642-35289-8_26}

\bibitem{deliege2020ordinal}
A.~Deli{\`e}ge, M.~Istasse, A.~Kumar, C.~De~Vleeschouwer, M.~Van~Droogenbroeck,
  Ordinal pooling, in: 30th British Machine Vision Conference, 2020.

\bibitem{Coates2011}
A.~Coates, A.~Arbor, A.~Y. Ng, {An Analysis of Single-Layer Networks in
  Unsupervised Feature Learning}, Aistats 2011 (2011) 215--223\href
  {http://arxiv.org/abs/fa} {\path{arXiv:fa}}, \href
  {http://dx.doi.org/10.1109/ICDAR.2011.95} {\path{doi:10.1109/ICDAR.2011.95}}.

\bibitem{Koniusz2013}
P.~Koniusz, F.~Yan, K.~Mikolajczyk,
  \href{http://www.sciencedirect.com/science/article/pii/S1077314212001725}{{Comparison
  of mid-level feature coding approaches and pooling strategies in visual
  concept detection}}, Computer Vision and Image Understanding 117~(5) (2013)
  479--492.
\newblock \href {http://dx.doi.org/10.1016/j.cviu.2012.10.010}
  {\path{doi:10.1016/j.cviu.2012.10.010}}.
\newline\urlprefix\url{http://www.sciencedirect.com/science/article/pii/S1077314212001725}

\bibitem{Boureau2010a}
Y.-L. Boureau, J.~Ponce, Y.~LeCun,
  \href{http://www.ece.duke.edu/{~}lcarin/icml2010b.pdf}{{A Theoretical
  Analysis of Feature Pooling in Visual Recognition}}, Icml (2010)
  111--118\href {http://dx.doi.org/citeulike-article-id:8496352}
  {\path{doi:citeulike-article-id:8496352}}.
\newline\urlprefix\url{http://www.ece.duke.edu/{~}lcarin/icml2010b.pdf}

\bibitem{IBPRIA2017}
M.~Pagola, J.~I. Forcen, E.~Barrenechea, J.~Fern{\'a}ndez, H.~Bustince, A study
  on the cardinality of ordered average pooling in visual recognition, in:
  L.~A. Alexandre, J.~Salvador~S{\'a}nchez, J.~M.~F. Rodrigues (Eds.), Pattern
  Recognition and Image Analysis, Springer International Publishing, Cham,
  2017, pp. 437--444.

\bibitem{Pagola2017}
M.~Pagola, J.~I. Forcen, E.~Barrenechea, C.~Lopez-Molina, H.~Bustince, Use of
  owa operators for feature aggregation in image classification, in: 2017 IEEE
  International Conference on Fuzzy Systems (FUZZ-IEEE), 2017, pp. 1--6.
\newblock \href {http://dx.doi.org/10.1109/FUZZ-IEEE.2017.8015606}
  {\path{doi:10.1109/FUZZ-IEEE.2017.8015606}}.

\bibitem{Scherer2010}
D.~Scherer, A.~M{\"u}ller, S.~Behnke, Evaluation of pooling operations in
  convolutional architectures for object recognition, in: K.~Diamantaras,
  W.~Duch, L.~S. Iliadis (Eds.), Artificial Neural Networks -- ICANN 2010,
  Springer Berlin Heidelberg, Berlin, Heidelberg, 2010, pp. 92--101.

\bibitem{Zeiler2013}
M.~D. Zeiler, R.~Fergus, \href{http://arxiv.org/abs/1301.3557}{{Stochastic
  Pooling for Regularization of Deep Convolutional Neural Networks}} (2013)
  1--9\href {http://arxiv.org/abs/1301.3557} {\path{arXiv:1301.3557}}.
\newline\urlprefix\url{http://arxiv.org/abs/1301.3557}

\bibitem{HUANG2014225}
Y.~Huang, Z.~Wu, L.~Wang, C.~Song,
  \href{http://www.sciencedirect.com/science/article/pii/S0925231213009636}{Multiple
  spatial pooling for visual object recognition}, Neurocomputing 129 (2014) 225
  -- 231.
\newblock \href
  {http://dx.doi.org/https://doi.org/10.1016/j.neucom.2013.09.037}
  {\path{doi:https://doi.org/10.1016/j.neucom.2013.09.037}}.
\newline\urlprefix\url{http://www.sciencedirect.com/science/article/pii/S0925231213009636}

\bibitem{ZHANG2018}
B.~Zhang, Q.~Zhao, W.~Feng, S.~Lyu,
  \href{http://www.sciencedirect.com/science/article/pii/S0925231218310610}{Alphamex:
  A smarter global pooling method for convolutional neural networks},
  Neurocomputing\href
  {http://dx.doi.org/https://doi.org/10.1016/j.neucom.2018.07.079}
  {\path{doi:https://doi.org/10.1016/j.neucom.2018.07.079}}.
\newline\urlprefix\url{http://www.sciencedirect.com/science/article/pii/S0925231218310610}

\bibitem{Cui2017}
Y.~Cui, F.~Zhou, J.~Wang, X.~Liu, Y.~Lin, S.~Belongie, Kernel pooling for
  convolutional neural networks, in: 2017 IEEE Conference on Computer Vision
  and Pattern Recognition (CVPR), 2017, pp. 3049--3058.
\newblock \href {http://dx.doi.org/10.1109/CVPR.2017.325}
  {\path{doi:10.1109/CVPR.2017.325}}.

\bibitem{pmlr-v51-lee16a}
C.-Y. Lee, P.~W. Gallagher, Z.~Tu,
  \href{http://proceedings.mlr.press/v51/lee16a.html}{Generalizing pooling
  functions in convolutional neural networks: Mixed, gated, and tree}, in:
  A.~Gretton, C.~C. Robert (Eds.), Proceedings of the 19th International
  Conference on Artificial Intelligence and Statistics, Vol.~51 of Proceedings
  of Machine Learning Research, PMLR, Cadiz, Spain, 2016, pp. 464--472.
\newline\urlprefix\url{http://proceedings.mlr.press/v51/lee16a.html}

\bibitem{SUN201796}
M.~Sun, Z.~Song, X.~Jiang, J.~Pan, Y.~Pang,
  \href{http://www.sciencedirect.com/science/article/pii/S0925231216312905}{Learning
  pooling for convolutional neural network}, Neurocomputing 224 (2017) 96 --
  104.
\newblock \href
  {http://dx.doi.org/https://doi.org/10.1016/j.neucom.2016.10.049}
  {\path{doi:https://doi.org/10.1016/j.neucom.2016.10.049}}.
\newline\urlprefix\url{http://www.sciencedirect.com/science/article/pii/S0925231216312905}

\bibitem{Lowe2004}
D.~G. Lowe, \href{http://portal.acm.org/citation.cfm?id=996342}{{Distinctive
  image features from scale invariant keypoints}}, International Journal of
  Computer Vision 60 (2004) 91--110.
\newblock \href
  {http://dx.doi.org/http://dx.doi.org/10.1023/B:VISI.0000029664.99615.94}
  {\path{doi:http://dx.doi.org/10.1023/B:VISI.0000029664.99615.94}}.
\newline\urlprefix\url{http://portal.acm.org/citation.cfm?id=996342}

\bibitem{N.Dalal2005}
B.~T. {N. Dalal}, {Histograms of Oriented Gradients for Human Detection.}, IEEE
  Conference on Computer Vision and Pattern Recognition (CVPR).

\bibitem{Cortes1995}
C.~Cortes, V.~Vapnik, {Support-Vector Networks}, Machine Learning 20~(3) (1995)
  273--297.
\newblock \href {http://arxiv.org/abs/arXiv:1011.1669v3}
  {\path{arXiv:arXiv:1011.1669v3}}, \href
  {http://dx.doi.org/10.1023/A:1022627411411}
  {\path{doi:10.1023/A:1022627411411}}.

\bibitem{Yager1988}
R.~Yager, \href{http://dl.acm.org/citation.cfm?id=46931.46950}{{On ordered
  weighted averaging aggregation operators in multi criteria decision making}},
  IEEE Trans. Syst. Man Cybern. 18~(1) (1988) 183--190.
\newblock \href {http://dx.doi.org/10.1109/21.87068}
  {\path{doi:10.1109/21.87068}}.
\newline\urlprefix\url{http://dl.acm.org/citation.cfm?id=46931.46950}

\bibitem{beliakov2016practical}
G.~Beliakov, H.~B. Sola, T.~C. S{\'a}nchez, A Practical Guide to Averaging
  Functions, Springer, 2016.

\bibitem{5206757}
{Jianchao Yang}, {Kai Yu}, {Yihong Gong}, T.~{Huang}, Linear spatial pyramid
  matching using sparse coding for image classification, in: 2009 IEEE
  Conference on Computer Vision and Pattern Recognition, 2009, pp. 1794--1801.
\newblock \href {http://dx.doi.org/10.1109/CVPR.2009.5206757}
  {\path{doi:10.1109/CVPR.2009.5206757}}.

\bibitem{Krizhevsky09}
A.~Krizhevsky, Learning multiple layers of features from tiny images, Tech.
  rep. (2009).

\bibitem{zhong2020random}
Z.~Zhong, L.~Zheng, G.~Kang, S.~Li, Y.~Yang, Random erasing data augmentation,
  in: Proceedings of the AAAI Conference on Artificial Intelligence (AAAI),
  2020.

\bibitem{NIPS2018_8271}
G.~Ghiasi, T.-Y. Lin, Q.~V. Le,
  \href{http://papers.nips.cc/paper/8271-dropblock-a-regularization-method-for-convolutional-networks.pdf}{Dropblock:
  A regularization method for convolutional networks}, in: S.~Bengio,
  H.~Wallach, H.~Larochelle, K.~Grauman, N.~Cesa-Bianchi, R.~Garnett (Eds.),
  Advances in Neural Information Processing Systems 31, Curran Associates,
  Inc., 2018, pp. 10727--10737.
\newline\urlprefix\url{http://papers.nips.cc/paper/8271-dropblock-a-regularization-method-for-convolutional-networks.pdf}

\end{thebibliography}

\end{document}